\documentclass[runningheads]{llncs}

\usepackage{eccv}

\usepackage{tikz}
\usepackage{pgfplots}
\usepackage{rotating}

\usepackage{amsmath}
\usepackage{soul}
\usepackage{bbding}
\usepackage{multirow}

\newcommand{\acro}{\textit{Valeo4Cast}}
\newcommand{\mapf}{mAP$_\text{f}$}
\newcommand*\samethanks[1][\value{footnote}]{\footnotemark[#1]}

\newcolumntype{a}{>{\columncolor{black!3}\color{black!50}}c}

\usepackage{makecell}
\definecolor{nicegreen}{RGB}{102, 204, 102}

\usepackage{eccvabbrv}

\usepackage{graphicx}
\usepackage{booktabs}

\usepackage[accsupp]{axessibility}  %

\usepackage{hyperref}
\usepackage[capitalize]{cleveref}
\crefname{section}{Sec.}{Secs.}
\Crefname{section}{Section}{Sections}
\Crefname{table}{Table}{Tables}
\crefname{table}{Tab.}{Tabs.}

\usepackage{orcidlink}

\begin{document}

\title{\acro{}: A Modular Approach to End-to-End Forecasting}

\author{
Yihong Xu\thanks{Core contributors. \\Correspondence to \tt\small {yihong.xu@valeo.com}}$\hspace{0.15cm}^1$ \and
\'Eloi Zablocki\samethanks$\hspace{0.15cm}^1$ \and
Alexandre Boulch\samethanks$\hspace{0.15cm}^1$ \and
Gilles Puy$^1$ \and
Mickael Chen$^1$ \and
Florent Bartoccioni$^1$ \and
Nermin Samet$^1$ \and
Oriane Siméoni$^1$ \and
Spyros Gidaris$^1$ \and
Tuan-Hung Vu$^1$ \and
Andrei Bursuc$^1$ \and
Eduardo Valle$^1$ \and
Renaud Marlet$^{1,2}$ \and
Matthieu Cord$^{1,3}$ \\[0.2cm]
}

\authorrunning{Xu et al.}

\institute{Valeo.ai, Paris, France \and
LIGM, Ecole des Ponts, Univ Gustave Eiffel, CNRS, Marne-la-Vallée, France \and
Sorbonne Universit\'e, Paris, France\\
\email{\{firstname,lastname\}@valeo.com}}

\maketitle

\begin{abstract}
  Motion forecasting is crucial in autonomous driving systems to anticipate the future trajectories of surrounding agents such as pedestrians, vehicles, and traffic signals. In end-to-end forecasting, the model must jointly detect and track from sensor data (cameras or LiDARs) the past trajectories of the different elements of the scene and predict their future locations. We depart from the current trend of tackling this task via end-to-end training from perception to forecasting, and instead use a modular approach. We individually build and train detection, tracking and forecasting modules. We then only use consecutive finetuning steps to integrate the modules better and alleviate compounding errors. We conduct an in-depth study on the finetuning strategies and it reveals that our simple yet effective approach significantly improves performance on the end-to-end forecasting benchmark. Consequently, our solution ranks first in the Argoverse 2 End-to-end Forecasting Challenge, with 63.82 \mapf{}. We surpass forecasting results by +17.1 points over last year's winner and by +13.3 points over this year's runner-up. This remarkable performance in forecasting can be explained by our modular paradigm, which integrates finetuning strategies and significantly outperforms the end-to-end-trained counterparts.
  \keywords{Finetuning \and  End-to-end motion forecasting \and Modular approach}
\end{abstract}

\section{Introduction}

{
Autonomous and assisted driving requires an accurate understanding of the scene surrounding the vehicle.
In particular, detecting \cite{liu2023bevfusion,chen2023focaformer3D,hu2023ealss,pointformer,peri2022lt3d}, tracking \cite{yin2021centerpoint,weng2020ab3dmot,xu2023transcenter} and forecasting \cite{salzmann2020trajectron++,benyounes2022cab,nayakanti2023wayformer,girgis2022autobot,shi2022mtr} the behavior of the agents (i.e., objects) in the scene, {agents which might be}  static {or} dynamic, is needed to plan the trajectory of the ego vehicle.}

{In recent years, these tasks have been tackled conjointly in pipelines that perform detection, tracking, and forecasting, as part of the same integrated network trained end-to-end, with great success \cite{wang2023le3de2e,peri22futuredet}. {We name such methods \emph{end-to-end-trained}.}}
In particular, ViP3D \cite{gu2023vip3d} introduced an end-to-end training pipeline from detection, tracking, and mapping to forecasting, and UniAD \cite{hu2023uniad} improved the forecasting performance and extended the pipeline to planning.

Despite these achievements, a recent study \cite{xu2024end2endforecast} reveals that current state-of-the-art end-to-end-trained approaches \cite{gu2023vip3d, hu2023uniad} are not without issues.
Crucially, {it shows} that a simple baseline that assembles independently trained detection, tracking, and forecasting modules outperforms the end-to-end training in the final forecasting task. {However,} because the modules {of this simple pipeline} are trained in isolation using curated data, {the} errors {of the} early modules are not compensated downstream, {which can} lead to {dramatic} compounding errors at the end of the pipeline.

{Following the findings of \cite{xu2024end2endforecast}, we {focus on advancing the forecasting performance} and build {in this work} a modular approach ({illustrated in \autoref{fig:archi}}). 
\begin{figure}[t]
    \centering
    \includegraphics[trim={0cm 1.9cm 1.2cm 0.7cm},clip,width=\linewidth]{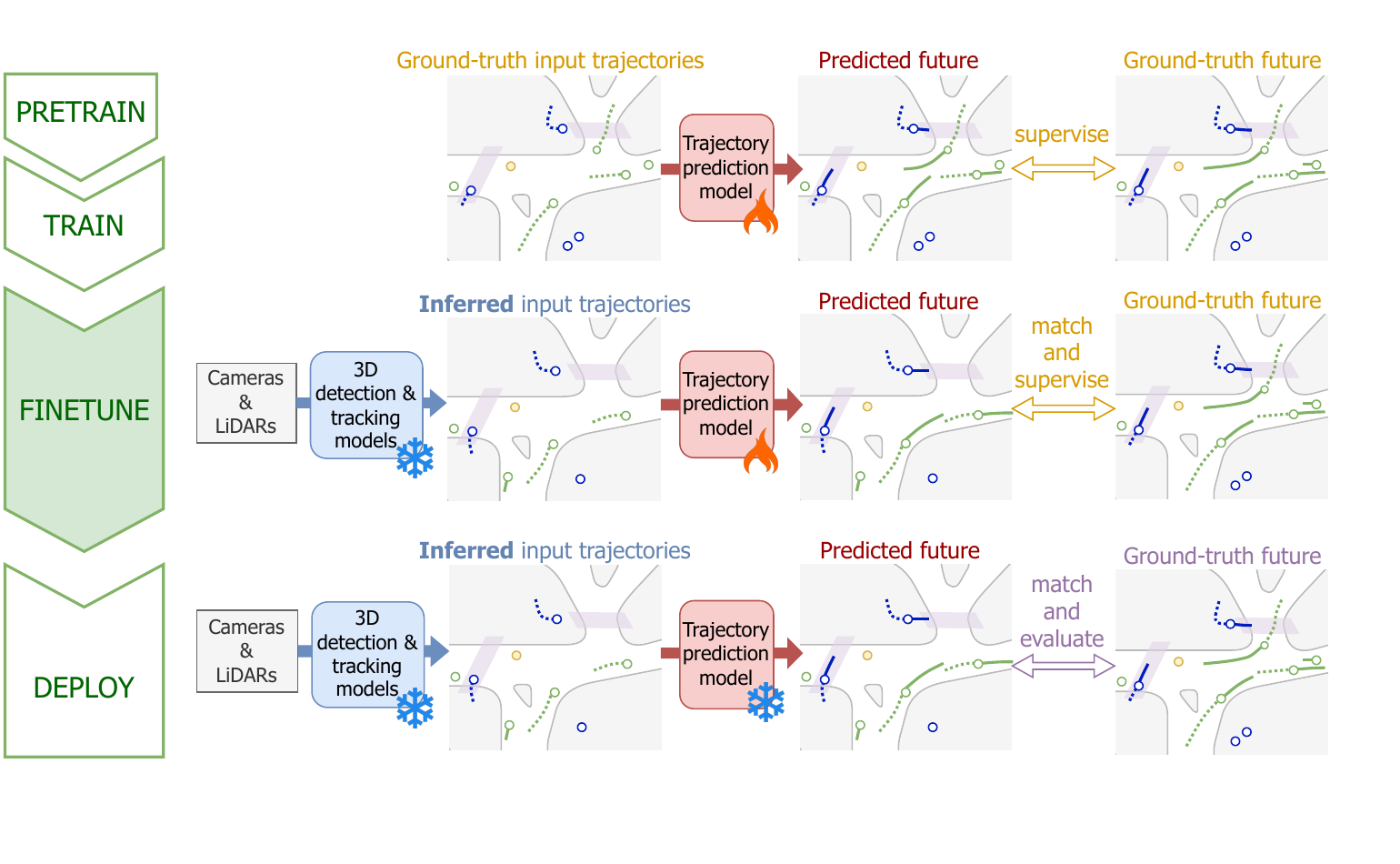}
    \caption{\textbf{Overview of the modular approach of \acro{}}. Conventional motion forecasting benchmarks provide curated annotations of past trajectories. Differently in this `end-to-end forecasting' challenge, we opt for a modular approach where the past trajectories are predicted by the detection and tracking modules. The predicted results contain imperfections such as FPs, FNs, IDS, and localization errors, which hinder forecasting. For this reason, training only on curated data is not sufficient (top). We thus propose a finetuning strategy where we match the predicted results and ground-truth annotations. We finetune the model on the matched pairs (middle) and it shows significant improvements once the model is deployed in real-world end-to-end forecasting (bottom). The \textcolor{orange}{ego car}, \textcolor{OliveGreen}{vehicles}, and \textcolor{blue}{pedestrians} are expressed in different colors. The past trajectories are shown with dotted lines and the future ones with plain lines. `Pretrain' refers to the pretraining on the UniTraj \cite{feng2024unitraj} framework, and `Train' to the step where we keep training on the curated Argoverse2-Sensor dataset.
    }
    \label{fig:archi}
\end{figure}
In particular, we use BEVFusion \cite{liu2023bevfusion} for detection, AB3DMOT \cite{weng2020ab3dmot} for tracking, and MTR \cite{shi2022mtr} for forecasting, and work on integrating all three into an end-to-end forecasting pipeline.
We start by pretraining the detection and forecasting modules individually with data curated for their respective tasks, 
the tracker has no trainable parameters.
To mitigate the compounding errors, we then finetune the forecasting module, using as input the outputs of the previous blocks. We observe in this challenge the importance of this adaptation step which drastically boosts performance.
Overall, this modular approach has the benefit to
(1) require limited resources as each functional block {is trained separately}  --- which is not the case for end-to-end training pipelines. 
It also (2) greatly improves the performances of the downstream blocks and
(3) opens the possibility of updating/upgrading a block without retraining all the upstream components.
The proposed pipeline is evaluated on the Argoverse Sensor forecasting benchmark \cite{wilson2021argoverse2} in the end-to-end forecasting paradigm.

We summarize here the main findings of the study, which are later discussed:

\begin{itemize}
    \item \textbf{Pretraining} {it} on a large dataset {helps better initialize the model}; 
    \item \textbf{Finetuning} {the forecasting module}
    on {predicted} {detection and tracking} inputs helps {to} take into account the errors of the {previous} detection and tracking blocks. We also reveal the impact of the forecasting performance with different finetuning strategies;
    \item \textbf{Post-processing} is then needed to ensure a valid trajectory for static objects.
\end{itemize}

This paper is organized as follows:
we recapitulate in~\autoref{sec:relatedwork} the related works of the perception pipeline and motion forecasting. We summarize in~\autoref{subsec:perception} the used perception models that generate detection and tracking results. We detail in~\autoref{subsec:forecasting} the forecasting model and our pretraining, finetuning, and post-processing strategies. In~\autoref{sec:experiments}, we compare with other competitors on the Argoverse 2 Sensor end-to-end forecasting benchmark and we conduct ablations such as the study of different finetuning strategies, which inspire the community towards a more robust end-to-end motion forecasting solution. 

\section{Related work}
\label{sec:relatedwork}
\paragraph{Perception pipeline.}
In autonomous driving, motion forecasting takes the results of detection and tracking as inputs. In terms of detection, the objective is to find the positions of objects surrounding the ego vehicle. The detectors can be grouped into three main modalities: Camera-only, LiDAR-only, and Camera-LiDAR. For models using cameras \cite{zhang2022mutr3d, gu2023vip3d, hu2023uniad}, the (RGB) images captured from cameras surrounding the ego vehicle are input to the network, and often projected into a bird's-eye-view representation for decoding. Although the images from the camera can provide more semantic information, such as the color of objects and fine-grained shape of (small) objects, camera-only methods suffer from a limited range of view. To this end, LiDAR information provides a longer detection range and 3D geometry of objects, which helps LiDAR-based or Camera-LiDAR-based detectors \cite{yin2021centerpoint, chen2023voxelnext, liu2023bevfusion} yield a better detection performance.  The detection results are then processed by a tracking algorithm that associates them over time. Among them, \cite{weng2020ab3dmot} is often leveraged to perform a per-class association in a non-differentiable way based on geometry cues such as intersection-over-union (IoU). Differently, recent end-to-end trained methods \cite{gu2023vip3d, hu2023uniad, xu2023transcenter} train and perform detection and tracking in a single network, resulting in a more efficient perception pipeline. However, they currently still underperform modular approaches. 

\paragraph{Motion forecasting.} Surprisingly, the conventional motion forecasting methods \cite{salzmann2020trajectron++, yuan2021agentformer, kim2021lapred, nayakanti2023wayformer, shi2022mtr} take instead the human-annotated detection and tracking ground-truth results as inputs. Ignoring the imperfect perception predictions, these works mainly focus on how to leverage ground-truth perception inputs by 1) interacting with the scenes (traffic light, traffic signs, roads, etc) and neighboring agents (surrounding vehicles, pedestrians, and other objects), 2) encoding the trajectory history of agents of interest, i.e., vehicles to forecast. This is often achieved by multiple attention modules where different information correlates with each other \cite{shi2022mtr, nayakanti2023wayformer}. 
With this information, a prediction head forecasts several future trajectories and a classification head gives the probabilities of each trajectory. Notably, recent anchor-based forecasters \cite{shi2022mtr} achieve state-of-the-art performance, mainly thanks to the pre-computed anchors, i.e., end locations, based on the distribution of the dataset. The anchors provide more diverse predicted trajectories, which also ease the probability prediction from the classification head.

Differently, in the Argoverse2 end-to-end motion forecasting challenge, we forecast the future trajectories based on predicted perception results at each time step, which is closer to the real-world deployment of a forecasting method. It is a more challenging task due to imperfect perception predictions, such as false or miss detections, identity switching, etc. In this track, recent end-to-end trained methods \cite{gu2023vip3d, hu2023uniad} enable the possibility of training perception tasks and forecasting in a joint manner. Although exciting as the single pipeline is an important step towards real-time forecasting, authors in \cite{xu2024end2endforecast} reveal that current end-to-end-trained forecasting methods do not exhibit on-par performance compared to a modular cascade of perception and forecasting modules, due to sub-optimized training and ignorance of map information. Based on these observations, we opt for a modular solution that is flexible to plug and play, and that significantly improves the end-to-end forecasting performance. Moreover, from our best knowledge, we conduct for the first time an in-depth study on the way we adapt the imperfect real-world perception results to the motion forecaster pretrained on ground-truth inputs.

\section{{Our approach}}
\label{sec:method}

We use in this work the modular pipeline represented in \autoref{fig:archi}. It consists of three independent modules for detection, tracking, and forecasting. We describe the perception modules and the forecasting method in the following subsections. We also provide implementation details for each block.

\subsection{Perception}
\label{subsec:perception}

{We first discuss our detection module (\autoref{subsec:detection}), followed by our tracking block (\autoref{subsec:tracking}).}

\subsubsection{Detection}
\label{subsec:detection}

\paragraph{Detection backbone.}
We use the LiDAR-only detector of BEVFusion \cite{liu2023bevfusion}.
It is composed of a sparse convolutional encoder which produces BEV features followed by a convolutional BEV backbone and multiple task-specific heads.
These heads predict the box center, dimension, and orientation of all objects.

\paragraph{Implementation details.}
We use a voxel size of 0.075~m and a BEV cell size of 0.6~m. As an addition to the current frame, we load the 5 previous lidar sweeps to densify the point cloud.
We train three different models: one detector that works in a range of up to 54~m}, another working in a range of up to 75~m range, and a third working in a range of up to 54~m but where the input features are enriched with ScaLR~\cite{puy2023revisiting} point features. These detectors are trained using up to 8 NVIDIA RTX 2080 Ti.

\paragraph{Ensembling.}
We then use a very simple heuristic to combine the detection of these three models.
For each timestep and each predicted class category, given a reference detection, we combine all detections with centers at a distance less than $r$ from the reference center.
We proceed greedily and consider the boxes by decreasing confidence order, removing the merged detection from the pool of detection to be processed.
The merging then consists of a simple weighted average, with the weights based on the boxes' center confidence, dimensions, and orientations.

\paragraph{Going further.}
As our focus {here} was on forecasting,  we used a simple  {LiDAR-based} detector {and trained it on only 10\% of the train set, with a limited range of 75m}. Perspectives {for this work} include {training the model on} all annotations, {and also leveraging} the cameras and the map information available in the dataset{, which could help produce stronger perception results and therefore improve the downstream forecasting}.

\subsubsection{Tracking}
\label{subsec:tracking}
\paragraph{{Tracking algorithm.}}
We adopt {the} simple and effective training-free tracking algorithm {of} AB3DMOT \cite{weng2020ab3dmot} to associate the detection results obtained in different frames. Precisely, we perform per-class tracking by running a one-to-one matching algorithm based on the track (object being tracked)-detection distances. The distance is determined by the 3D intersection-over-union (IoU) between tracks and detection at each time step, and the matching threshold is set to 0.1 for all classes. Moreover, a track may be temporarily lost due to occlusions. In this case, this track is put into `inactive' mode, and its position is updated 
with a Kalman filter until it is matched {to} a detection. The `inactive' mode can only last 3 frames, after which the track is terminated.

\paragraph{Linear interpolation of tracks.}
{When using the tracking algorithm presented above,} we observe {that the trajectories can be fragmented into sub-trajectories.}
{To mitigate the problem}, inspired by ByteTrack \cite{Bytetrack}, we linearly interpolate {between the fragment} trajectories  {using} a constant velocity calculated {using} the object locations at current and past timesteps. This interpolation improves HOTA by 0.45 points. \cite{HOTA}.
The general tracking approach forgoes any training.

\subsection{Forecasting}
\label{subsec:forecasting}
{To forecast the different agents,} we use the MTR \cite{shi2022mtr} forecasting model, which won the 2022 Waymo forecasting challenge. The architecture is transformer-based {and has} 60M trainable parameters. It jointly learns the agent's global intentions as well as their local movements. 

\paragraph{Pretraining.}
We pretrain MTR on 1300$+$ hours of vehicle trajectories with the UniTraj framework that gathers nuScenes \cite{caesar2020nuscenes}, Argoverse2-Motion \cite{wilson2021argoverse2}, and Waymo Open Motion Dataset (WOMD) \cite{ettinger2021waymo}.
{We then {further} train MTR on the curated Argoverse2-Sensor dataset.}

\paragraph{Finetuning.}
At inference time, the forecasting model is applied to the outputs of the perceptions modules (detection and tracking), which are imperfect, with misdetections, hallucinations, tracking issues, and localization errors.
{To deal with such mistakes,} we finetune MTR on such  {imperfect} data.
In practice, given a track predicted by our upstream perception modules, we match it with the ground-truth trajectory provided in {Argoverse 2}  training annotations {in order} to get the future ground-truth to predict. 

{Since the detections are not filtered by any detection score, they are {typically} redundant.
To provide rich supervision {for the} forecasting, we perform a many (predictions)-to-one (ground truth) matching based on the Euclidean distance between the past trajectories of the tracks and the ground-truth annotations at each inference time step. {In} the distance calculation, we only consider the past timestamps where the prediction and ground truth are both available. For the matched track and ground truth, we train MTR to predict the corresponding ground-truth future trajectory given the matched predicted past trajectory. We finetune the model only on the trainset for 15 epochs and choose the checkpoint with the lowest brier-FDE on the validation set. {Given our results, discussed in} \autoref{tab:forecastingAblation}, Different finetuning strategies are studied in the following section, and the chosen finetuning strategy appears crucial for the forecasting performance.} 

\paragraph{Post-processing.}
{Because the pretraining was performed on standard (non-end-to-end) forecasting data that do not contain static trajectories, our model tends to avoid predicting static motion. This {can} be easily solved {using a} post-processing {step}.}
During inference, we conduct the following steps:
\begin{itemize}
    \item Static trajectories are the most prevalent {in the dataset}, {however the forecasting module is trained mainly on \emph{moving objects}.} We {therefore} insert a static trajectory {in the predictions}: {we}  replace the least probable mode with a stationary future, and assign it a score of 1.
    \item For object classes that are always static\footnote{These include 'bollard', 'construction cone', 'construction barrel', 'sign', 'mobile pedestrian crossing sign', and 'message board trailer'}, we predict a single static trajectory with a probability of 1. This only marginally impacts the scores.
\end{itemize}
As the \mapf{} computation is performed at the trajectory type level and then averaged, we find that both steps significantly boost the score, as seen in \autoref{tab:forecastingAblation} (`without post-processing line). 
{The post-processing} {currently requires no training, but could be improved by predicting if a future trajectory is likely to be static or not.}

\paragraph{Implementation details.}
We build our work based on the implementation of UniTraj~\cite{feng2024unitraj}.
We use the past 2 seconds (or until the beginning of the sequence), even though we are allowed to use all past frames. {The finetuning takes around 12 hrs on a single node with {8$\times$A100 GPUs}.}

\section{Experiments and Results}
\label{sec:experiments}

\subsection{Dataset}
We train and evaluate our method on the Argoverse 2 Sensor Dataset \cite{wilson2021argoverse2}.
{It contains} 4.2 hours of driving data, split into 1000 scenes (750/150/150 for train/val/test).
Each scene lasts about 15 seconds with sensor data and annotations being provided at a 10Hz sampling rate.
{The input data include images captured from a 360°-rig of 7 cameras, LiDAR scans, and HD-maps of the environment that includes information about lines, drivable areas and crosswalks.}
Annotations are provided for 26 different semantic classes, including common ones like vehicle and bus and less frequent ones like wheelchair, construction cone, dog, and message board trailer.

\subsection{Evaluation metrics}
\paragraph{Detection.}
The detection performance is measured with the mCDS metric. 
The Composite Detection Score (CDS) gathers in a single score the detection precision, recall, and quality of the estimation of the object extent, positioning, and orientation.
This metric is averaged over all object classes to form mCDS.
The evaluation range is 150m around the ego-vehicle.

\paragraph{Tracking.}
{HOTA \cite{HOTA} is the main metric used for evaluating the tracking performance. It breaks down the detection and association evaluation by calculating separately the False Positives (FP), False Negatives (FN), and True Positives (TP). It alleviates the issue of overly emphasizing detection performance in the multi-object tracking accuracy (MOTA) \cite{clearmot} metric, which calculates the sum of FP, FN, and IDentity Switch (IDS) over the total number of ground-truth objects. MOTA reflects the overall performance of a multi-object tracker with a focus on detection. Since MOTA only considers the tracking result after thresholding, a variant of MOTA -- AMOTA averaging all recall thresholds \cite{weng2020ab3dmot}. These metrics are averaged over all object classes.
The evaluation range is 50m around the ego vehicle.}

\paragraph{Forecasting.}
The challenge's primary metric is the mean Forecasting Average Precision (\mapf{}) \cite{peri22futuredet}. 
This metric shares the same formulation as detection AP \cite{lin2014microsoft}. 
However, in the case of \mapf{}, a true positive is defined for trajectories that match at the current time step (that is, agents successfully detected) and the final time step (that is, agents successfully forecasted).
For agents that are successfully detected, the other metrics considered are average displacement error (ADE) and final displacement error (FDE), where ADE measures the average Euclidean distance between the predicted and ground-truth positions over the future time horizon, while FDE measures the distance at the final time step. {We stress that ADE and FDE can only compute a distance when a ground truth and a predicted detection have been matched and therefore do not account for miss-detected or hallucinated agents. In fact, not detecting the more difficult agents can improve ADE and FDE errors.
Therefore, these metrics should be used carefully in the context of end-to-end forecasting benchmarks to avoid erroneous interpretations.}
The evaluation range is 50m around the ego-vehicle.

\subsection{Results and discussion}

\paragraph{{Leaderboard results.}}
{We provide the leaderboard results from Argoverse 2 end-to-end forecasting challenge in~\autoref{tab:leaderboardresults}.} {The proposed modular solution \acro\, achieves {strong} 
performance on forecasting, outperforming the second-best solution by more than 13 points on \mapf{}.}
This demonstrates the usefulness of our modular approach, {which} allows us to easily transfer the strong pretrained forecasting model such as MTR to the end-to-end task via a suitable finetuning strategy.
Interestingly, even though we depart from lower detection and tracking results compared to other current methods, our finetuned trajectory prediction model can overcome the difference and significantly outperform them. {From \autoref{tab:nonlinear}, we observe that \acro{} outperforms its competitors by large margins in predicting non-linear trajectories, with more than 50\% improvements in \mapf{} and ADE and FDE. This is made possible with our finetuning and post-processing strategies.}
\begin{table}[h]
\centering
\renewcommand{\arraystretch}{1.3}
\resizebox{\linewidth}{!}{%
\begin{tabular}{cl ccc aaa a}

\toprule
& & \multicolumn{3}{c}{\textbf{Forecasting}} & \multicolumn{3}{a}{\textbf{Detection}} & \multicolumn{1}{a}{\textbf{Tracking}}\\
\cmidrule(lr){3-5} \cmidrule(lr){6-8} \cmidrule(lr){9-9}
&  & \mapf{} ($\uparrow$) & ADE ($\downarrow$) & FDE ($\downarrow$) &
Inputs & Training range & mCDS ($\uparrow$) & HOTA ($\uparrow$)
\\
\midrule
\multirow{6}{*}{\rotatebox[origin=c]{90}{2023}}
& CenterPoint \cite{yin2021centerpoint} & - & - & - & L & 150 & 14 & -\\
& BEVFusion \cite{liu2023bevfusion} & - & - & - & L+C & 150 & 37 & - \\
& Anony\_3D (v0) \cite{lee2024revoxeldet} & - & - & - & L & 150 & 31 & 44.36 \\ %
& Host\_4626\_Team \cite{peri2022lt3d} & 14.51 & 5.10 & 7.32 & - & - & - & 39.98\\
& Dgist-cvlab \cite{woo2023motion} & 42.91 & 4.11 & 4.59 & L+C & 150 & 34 & 41.49\\
& Le3DE2E \cite{wang2023le3de2e} & 46.70 & 3.22 & 3.76 & L+C & 150 & 39 & 56.19\\
\midrule
& Dgist-cvlab & 45.83 & 4.09 & 4.53 & L+C & 150 & 34 & 41.49\\
& Le3DE2E & 50.53 & 4.07 & 4.60 & L+C & 150 & \textbf{43} & \textbf{64.60} \\
\multirow{-3}{*}{\rotatebox[origin=c]{90}{2024}} & \cellcolor{nicegreen!20!white}\acro{} & \cellcolor{nicegreen!20!white}\textbf{63.82}& \cellcolor{nicegreen!20!white}\textbf{2.14} & \cellcolor{nicegreen!20!white}\textbf{2.43} & L & 75 & 31 & 61.28\\
\bottomrule
\end{tabular}
}
\caption{\textbf{Leaderboard results.} Test set results of the Argoverse 2 end-to-end forecasting leaderboard, with three sub-challenges: forecasting, detection and tracking. We distinguish the detectors by their input modality with `L' for LiDAR and `C' for Camera. For all methods, the \textit{evaluation} range is fixed to 150~m for detection, and 50~m for tracking and forecasting. {Our modular strategy significantly outperforms others in the forecasting challenge (by more than 13 \mapf{} pts).}
}

\label{tab:leaderboardresults}
\end{table}

\begin{table}[h]
\renewcommand{\arraystretch}{1.3}
\centering
\begin{tabular}{@{}l cccc@{}}
\toprule
&   & \makecell{Non-linear \mapf{} ($\uparrow$)} & Non-linear ADE ($\downarrow$) & Non-linear FDE ($\downarrow$) \\
\midrule
Le3DE2E \cite{wang2023le3de2e} (2023) &  & 4.0& 7.6 & 8.7\\
Le3DE2E &  & 30.4& 10.7 & 11.8\\
\cellcolor{nicegreen!20!white}\acro{} & \cellcolor{nicegreen!20!white} & \cellcolor{nicegreen!20!white} \textbf{68.4} & \cellcolor{nicegreen!20!white} \textbf{4.0}& \cellcolor{nicegreen!20!white} \textbf{4.5}\\

\bottomrule
\end{tabular}
\caption{\textbf{Leaderboard results on non-linear trajectories.}  Test set results for non-linear trajectories of the Argoverse 2 end-to-end forecasting leaderboard.
}
\label{tab:nonlinear}
\vspace{-6mm}
\end{table}

\begin{table}[h]
\renewcommand{\arraystretch}{1.3}
\centering
\begin{tabular}{@{} cc|cc| cccc@{}}
\toprule
\multicolumn{2}{c|}{Assignment}& \multicolumn{2}{c|}{Distance} &  &  &   & \\
many-one &one-one  & $t \leq 0$ & $t=0$  & \mapf{} ($\uparrow$)  & \makecell{Non-linear\\\mapf{} ($\uparrow$)} & ADE ($\downarrow$) & FDE ($\downarrow$) \\
\midrule
& & & & 43.3 & 33.6& 1.97 &   3.24 \\
&\Checkmark & & \Checkmark& 62.1 & 64.7& 0.68 &   1.11 \\
&\Checkmark &\Checkmark & & 61.2& 60.2 & 0.68  &  1.12  \\
\Checkmark & & & \Checkmark& 62.6 & 65.0 & 0.66  &  0.99  \\
\cellcolor{nicegreen!20!white} \Checkmark &  \cellcolor{nicegreen!20!white}  &  \cellcolor{nicegreen!20!white} \Checkmark & \cellcolor{nicegreen!20!white} & \cellcolor{nicegreen!20!white}\textbf{63.0} & \cellcolor{nicegreen!20!white}\textbf{67.5} & \cellcolor{nicegreen!20!white}\textbf{0.60} & \cellcolor{nicegreen!20!white}\textbf{0.96}\\

\bottomrule
\end{tabular}
\caption{\textbf{Ablation study of the forecasting finetuning strategies.} Scores are reported on the validation set. The evaluation is conducted in a 50m-range around the ego-car. \mapf{} is the main metric of the challenge. {The distance threshold is set to 2 meters for assignment beyond which the prediction will not be assigned to any ground truth.}
}
\label{tab:finetuningAblation}
\end{table}

\begin{table}[ht]
\renewcommand{\arraystretch}{1.3}
\centering
\begin{tabular}{@{}l cccc@{}}
\toprule
& \mapf{} ($\uparrow$)  & \makecell{Non-linear\\\mapf{} ($\uparrow$)} & ADE ($\downarrow$) & FDE ($\downarrow$) \\
\midrule
\cellcolor{nicegreen!20!white}\acro{} & \cellcolor{nicegreen!20!white}\textbf{63.0} & \cellcolor{nicegreen!20!white}{67.5} & \cellcolor{nicegreen!20!white}\textbf{0.60} & \cellcolor{nicegreen!20!white}0.96\\
\makecell[l]{\hspace{0.5cm} w/o finetuning with\\\hspace{0.5cm}perception inputs} & 43.3  & 33.6     &  1.97 & 3.24\\
\hspace{0.5cm} w/o post-processing & 54.6 & 33.2 &  0.61 & \textbf{0.94}  \\
\hspace{0.5cm} w/o UniTraj pretraining & 62.9 & \textbf{68.3} & 0.62 & 0.98  \\
\bottomrule
\end{tabular}
\caption{\textbf{Ablation study of the forecasting module.} Scores are reported on the validation set. The evaluation is conducted in a 50m-range around the ego-car. \mapf{} is the main metric of the challenge.
}
\label{tab:forecastingAblation}
\end{table}

{\paragraph{Finetuning strategies.} In \cite{xu2024end2endforecast}, the authors reveal the significant drop from ground-truth to predicted inputs. Although they argue that finetuning with a simple one-to-one matching improves marginally performance, an in-depth study of how to finetune is still a missing brick. Therefore, we conduct the study on how to finetune a motion forecasting model with predicted inputs.

In the end-to-end forecasting setting, the forecasting module can no longer have access to ground-truth past trajectories during inference. Instead, these are produced by the perception pipeline on the fly: the detector detects objects in the past and the tracker associates the detection through time, which forms the estimated past trajectories. The predicted past trajectories are not by definition matched with the ground-truth trajectories. Therefore, to finetune the {forecasting module} \cite{shi2022mtr} with predicted past trajectories, a \emph{matching} procedure of the predicted and ground-truth past trajectories is needed. We remind that the {matching} strategies are only used for finetuning purposes, and the standard evaluation protocol remains unchanged. 
{
We examine four distinct finetuning approaches, each employing different methods for calculating the matching distance matrix and association. These methods are based on either the initial positions of the agents at the forecasting starting point ($t=0$) or all the previously accessible trajectory data ($t \leq 0$).
Additionally, we explore different assignment strategies, namely `many (predictions)-to-one' and `one-to-one':
}
\begin{itemize}
    \item one-to-one assignment, with distance at $t=0$.
    \item one-to-one assignment, with distance at $t<=0$.
    \item many-to-one assignment, with distance at $t=0$.
    \item many-to-one assignment, with distance at $t<=0$.

\end{itemize}
}
{
From \autoref{tab:finetuningAblation}, we can see that with many-to-one assignments, the overall performance is better than the one-to-one assignment. This is mainly because we do not perform any filtering strategy (e.g., filtering by scores) for the past trajectory, they are thus redundant. Performing many-to-one assignments can provide more training pairs for finetuning since several neighboring predictions will be matched to the same ground truth and be trained to predict similar future trajectories. Under this many-to-one assignment setting, matching the entire (available) past trajectories considers the driving direction of agents that matches with ground truth compared to only considering the distance at $t=0$, which provides more accurate training signals for finetuning and shows better forecasting performance after training.}

\begin{figure}[ht]
\centering
\begin{tikzpicture}
\begin{axis}[
    ybar,
    ymin=0,
    ymax=1,
    bar width=5pt,
    width=\textwidth,
    height=0.25\textheight,
    ylabel={\mapf{}},
    ylabel style={at={(axis description cs:0.05,0.5)},anchor=south},
    symbolic x coords={mob. ped. cross. sign, constr. barrel, constr. cone, reg. vehicle, bollard, motorcyclist, motorcycle, bicycle, box truck, truck, school bus, pedestrian, bus, stroller, wheeled rider, truck cab, bicyclist, stop sign, veh. trailer, wheeled dev., dog, large vehicle, sign, artic. bus, wheelchair},
    xtick=data,
    xticklabel style={rotate=45,anchor=east,font=\scriptsize},
]
\addplot coordinates {
    (mob. ped. cross. sign, 0.911)
    (constr. barrel, 0.909)
    (constr. cone, 0.814)
    (reg. vehicle, 0.802)
    (bollard, 0.788)
    (motorcyclist, 0.787)
    (motorcycle, 0.733)
    (bicycle, 0.719)
    (box truck, 0.718)
    (truck, 0.690)
    (school bus, 0.676)
    (pedestrian, 0.657)
    (bus, 0.656)
    (stroller, 0.649)
    (wheeled rider, 0.644)
    (truck cab, 0.555)
    (bicyclist, 0.545)
    (stop sign, 0.530)
    (veh. trailer, 0.522)
    (wheeled dev., 0.522)
    (dog, 0.514)
    (large vehicle, 0.497)
    (sign, 0.481)
    (artic. bus, 0.428)
    (wheelchair, 0.288)
};
\end{axis}
\end{tikzpicture}
\caption{\textbf{Per-class performance comparison of \acro{} with and without pretraining}. We show the per-class performance in \mapf{} of 26 classes in the Argoverse 2 sensor dataset for the end-to-end forecasting. The reported scores are on the validation set. The evaluation is conducted in a 50m-range around the ego-car.
\label{fig:per_class}
}
\end{figure}

\paragraph{{Ablation study.}} From \autoref{tab:forecastingAblation}, we observe that finetuning MTR on the trainset predictions of the used detector and tracker on the Argoverse2-Sensor data is crucial for forecasting performance. 
It improves drastically the \mapf{} from 43.3 to 63.0. This is mainly because the vanilla MTR model has never been trained on {predicted inputs.}
The finetuning can adapt the model not only to new trajectory types but also to the {inaccuracies in predicted inputs.}

{Surprisingly, after finetuning, we find that using the model pretrained on 1300+ hours of vehicle trajectories does not bring significant benefit compared to training from scratch (62.9 \mapf{}). We believe that it is due to the difference in object classes and the data distribution between ground truth and predicted past trajectories.}

{And finally, {compensating for the lack of static trajectories in pretraining}, the post-processing effectively helps to improve the forecasting performance.} Moreover, from \autoref{fig:per_class}, we observe that the forecasting performance of different object classes varies due to the imbalanced training examples in the dataset.
\begin{figure*}
    \centering
    \begin{subfigure}[b]{0.49\textwidth}
        \centering
        \includegraphics[width=\textwidth]{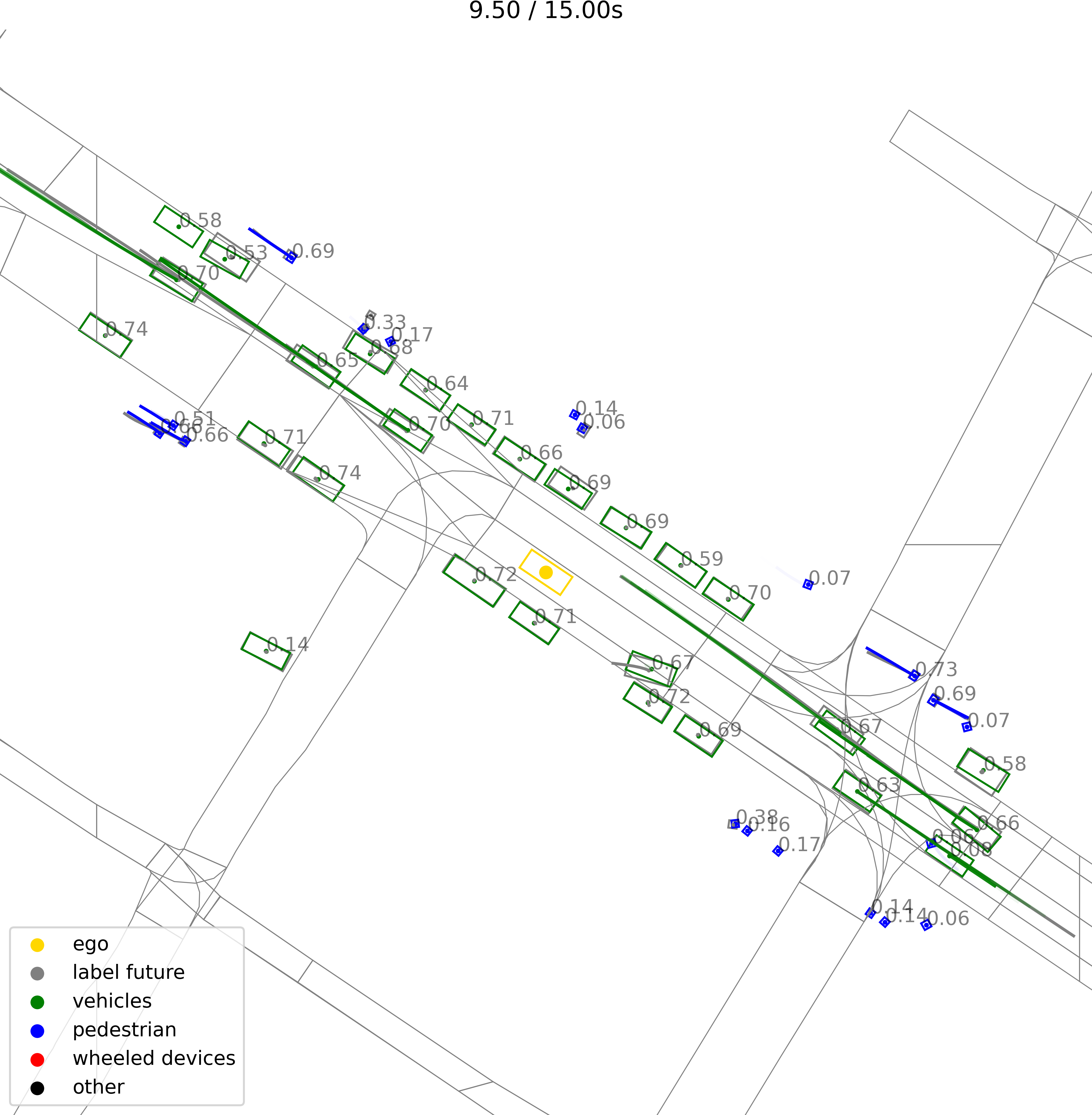}
    \end{subfigure}
    \hfill
    \begin{subfigure}[b]{0.49\textwidth}
        \centering
        \includegraphics[width=\textwidth]{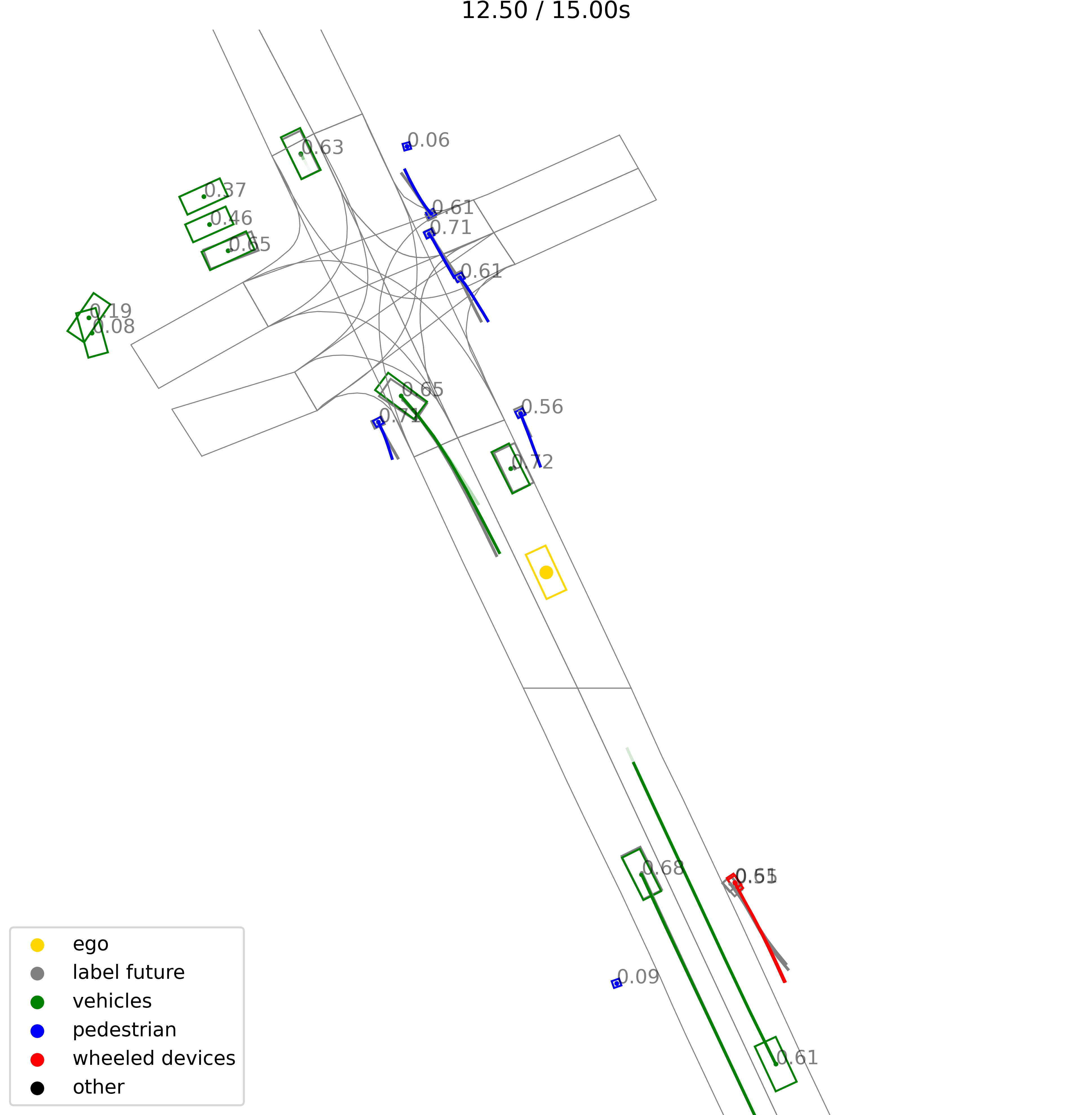}
    \end{subfigure}

    \begin{subfigure}[b]{0.49\textwidth}
        \centering
        \includegraphics[width=\textwidth]{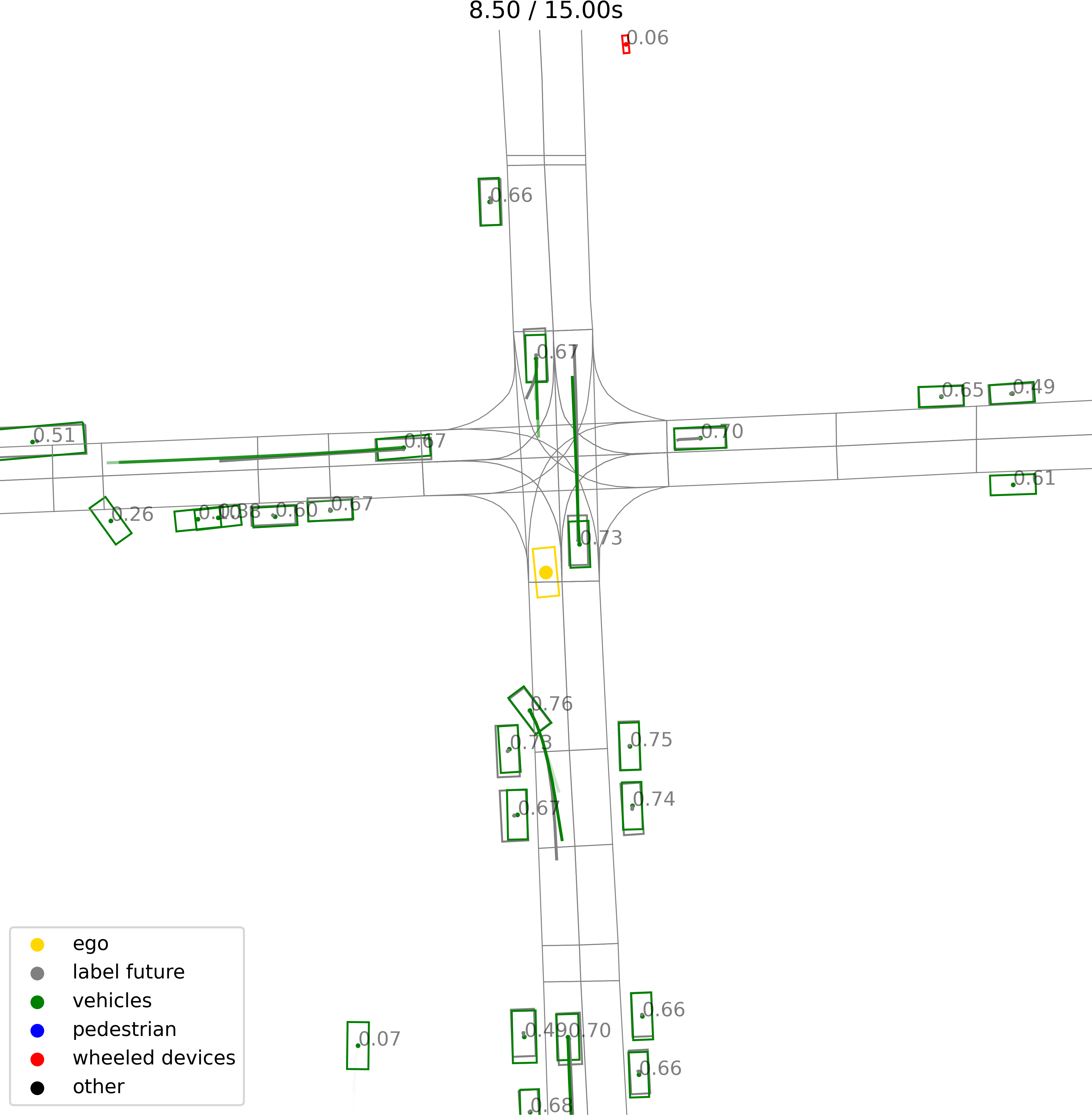}
    \end{subfigure}
    \hfill
     \begin{subfigure}[b]{0.49\textwidth}
        \centering
        \includegraphics[width=\textwidth]{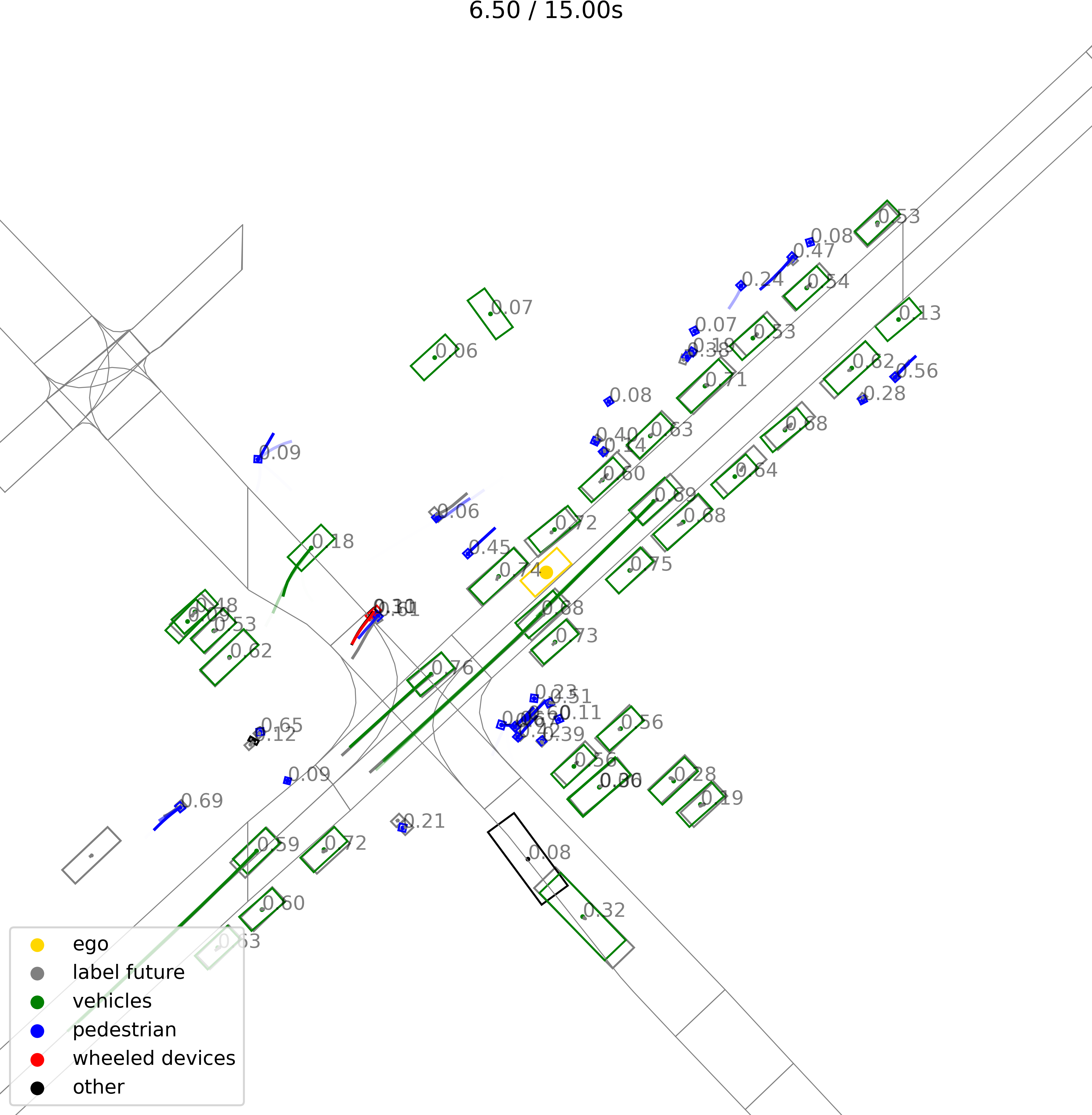}
    \end{subfigure}
    \caption{\textbf{Qualitative visualizations} of randomly sampled frames of the validation set. The \textcolor{Yellow!85!black}{ego car}, \textcolor{OliveGreen}{vehicles}, \textcolor{red}{wheeled devices}, \textcolor{blue}{pedestrians} and \textcolor{gray}{ground-truth annotations} are expressed in different colors. The numbers represent the detection scores.}
    \label{fig:main}
\end{figure*}
\paragraph{Visualizations} {We visualize examples of forecasting obtained on randomly selected frames in \autoref{fig:main}.} Despite some miss detection for distant objects, we observe that our perception predictions overlap well with the ground truth in gray, showing a reasonably good perception performance. As for forecasting, different modes cover well the possible future trajectories and the static trajectories from the post-processing ensure a good prediction for parked objects.

\section{Conclusion}
{The modular pipeline \acro\,ranks first in the AV2 E2E Forecasting challenge 2024 and outperforms other solutions by $+$13 \mapf{} pts.
{This design allowed us to start from a state-of-the-art motion forecasting model, that we integrate into an end-to-end pipeline by finetuning the forecasting module. We conduct in-depth study on how to finetune the forecaster with different matching strategies. We include a post-processing step to account for the absence of static objects in conventional motion forecasting pretraining which are important in the end-to-end benchmark.}
In this work, we {confirm the findings of \cite{xu2024end2endforecast}, verifying} the superiority of modular approaches in the end-to-end forecasting task, and their ability to handle detection and tracking imperfections.}

{The efficient nature of the end-to-end approaches is still appealing.} In future work, we are interested in investigating how to better train the end-to-end approaches to achieve performances on par with \acro.
Besides, future work may also consider more challenging settings in which the map information is not provided, at any stage, and has to be inferred in an online fashion, as the ego car drives and discovers its environment.

\bibliographystyle{splncs04}
\bibliography{main}
\end{document}